\documentclass[letterpaper, 10 pt, conference]{ieeeconf}  %

\IEEEoverridecommandlockouts                              %

\usepackage{cite}
\usepackage{amsmath,amssymb,amsfonts}
\usepackage{algorithmic}
\usepackage{graphicx}
\usepackage{textcomp}
\usepackage{xcolor}

\usepackage{comment}

\usepackage{xcolor}
\usepackage{pagecolor}
\usepackage{lipsum}  
\usepackage{mdframed}
\usepackage{tikz}
\usetikzlibrary{external}

\def\BibTeX{{\rm B\kern-.05em{\sc i\kern-.025em b}\kern-.08em
    T\kern-.1667em\lower.7ex\hbox{E}\kern-.125emX}}

\definecolor{TUMBlue}{RGB}{0,101,189}%
\definecolor{TUMWhite}{RGB}{255,255,255}%
\definecolor{TUMBlack}{RGB}{0,0,0}%
\definecolor{TUMBlue1}{RGB}{0,51,89}%
\definecolor{TUMBlue2}{RGB}{0,82,147}%
\definecolor{TUMGray1}{RGB}{51,51,51}%
\definecolor{TUMGray2}{RGB}{127,127,127}%
\definecolor{TUMGray3}{RGB}{204,204,204}%
\definecolor{TUMBlue3}{RGB}{100,160,200}%
\definecolor{TUMBlue4}{RGB}{152,198,234}%
\definecolor{TUMIvory}{RGB}{218,215,203}%
\definecolor{TUMOrange}{RGB}{227,114,34}%
\definecolor{TUMGreen}{RGB}{162,173,0}%

\usepackage{booktabs}
\usepackage[per-mode=fraction, detect-weight=true]{siunitx}
\usepackage[hidelinks]{hyperref}

\newcounter{subfigure}[figure]

\newcommand{\subfigcaptioncustom}[1]{
    \par\footnotesize(\alph{subfigure})\;
    #1
}
\newcommand{\startsubfigcustom}{\refstepcounter{subfigure}}
\newenvironment{subfigurecustom}
{%
    \refstepcounter{figure}%
    \setcounter{subfigure}{0}%
}
{%
    \addtocounter{figure}{-1}%
} %

\DeclareSIUnit\bar{bar}

\makeatletter
\def\fps@figure{t}
\def\fps@table{t}
\makeatother

\RequirePackage{letltxmacro}
\LetLtxMacro{\origfigure}{\figure}
\LetLtxMacro{\endorigfigure}{\endfigure}
\renewenvironment{figure}[1][]
{\origfigure[#1]\centering}
{\endorigfigure}

\LetLtxMacro{\origtable}{\table}
\LetLtxMacro{\endorigtable}{\endtable}
\renewenvironment{table}[1][]
{\origtable[#1]\centering}
{\endorigtable}

\newcommand{\inputtikzfig}[1]{
    \includegraphics{./images/#1.pdf}
}
\newcommand{\inputtikzplot}[1]{
    \includegraphics{./plots/#1.pdf}
}

\newcommand{\softwareshort}[1]{\textsc{#1}}
\newcommand{\softwarefull}[1]{#1}
\newcommand{\rslcppfull}{\softwarefull{ROS Simulation Library for C++}}
\newcommand{\rslcpp}{\softwareshort{rslcpp}}
\newcommand{\rclcpp}{\softwareshort{rclcpp}}
\newcommand{\rclc}{\softwareshort{rclc}}
\newcommand{\ros}{\softwareshort{ROS~2}}
\newcommand{\roswithouttwo}{\softwareshort{ROS}}
\newcommand{\kissicp}{\softwareshort{KISS\nobreakdash-ICP}}
\newcommand{\kissicpbreaking}{\softwareshort{KISS-ICP}}
\newcommand{\gazebo}{\softwareshort{Gazebo}}
\newcommand{\microros}{\softwareshort{micro\nobreakdash-ROS}}
\newcommand{\evo}{\softwareshort{evo}}
\newcommand{\kitti}{\softwareshort{KITTI}}
\newcommand{\kittiodometry}{\softwareshort{KITTI Odometry}}
\newcommand{\kittitobag}{\softwareshort{kitti2bag}}

\newcommand{\schedChartlabelTtwo}{$\mathcal{T}_2$}
\newcommand{\schedChartlabelTone}{$\mathcal{T}_1$}
\newcommand{\schedChartlabelSone}{$\mathcal{S}_1$} %
\usepackage{booktabs}
\usepackage[per-mode=fraction, detect-weight=true]{siunitx}
\usepackage[hidelinks]{hyperref}
 
\title{\LARGE \bf RSLCPP - Deterministic Simulations Using \ros{}}

\author{Simon Sagmeister$^{1}$, Marcel Weinmann$^{1}$, Phillip Pitschi$^{2}$, and Markus Lienkamp$^{1}$%
\thanks{*This work was funded by the Deutsche Forschungsgemeinschaft (DFG, German Research Foundation) – 469341384}%
\thanks{$^{1}$Technical University of Munich, Germany; School of Engineering \& Design, Department of Mobility Systems Engineering, Institute of Automotive Technology}
\thanks{$^{2}$Technical University of Munich, Germany; School of Engineering \& Design, Department of Engineering Physics and Computation, Institute of Automatic Control\newline{}
Corresponding author: \href{mailto:simon.sagmeister@tum.de}{simon.sagmeister@tum.de}}
}

\newcommand\copyrighttext{%
	\footnotesize \centering This is the accepted manuscript of an article published in IEEE Robotics and Automation Practice (RA-P).\\The version of record is available at \url{https://doi.org/10.1109/RAP.2026.3704080}.
}

\newcommand\copyrightnotice{%
	\tikzset{external/export=false}
	\begin{tikzpicture}[remember picture,overlay]
	\node[anchor=south,yshift=10pt, xshift=0pt] at (current page.south) {\fbox{\parbox{\dimexpr\textwidth-\fboxsep-\fboxrule\relax}{\copyrighttext}}};
	\end{tikzpicture}%
	\tikzset{external/export=true}
    \vspace{-0.35cm}
}

\begin{document}

\bstctlcite{IEEEexample:BSTcontrol}

\maketitle
\thispagestyle{empty}
\pagestyle{empty}
\copyrightnotice{}

\setcounter{footnote}{0}
\begin{abstract}
    Simulation is crucial in real-world robotics, offering safe, scalable, and efficient environments for developing a variety of robotic applications.
    While the \softwarefull{Robot Operating System} (\roswithouttwo{}) has been widely adopted as the backbone of these robotic applications in both academia and industry, its asynchronous, multi-process design complicates reproducibility, especially across varying hardware platforms. Deterministic callback execution cannot be guaranteed when computation times and communication delays vary.
    This lack of reproducibility complicates scientific benchmarking and continuous integration, where consistent results are essential.
    To address this, we present a methodology to create deterministic simulations using \ros{} nodes. Our \rslcppfull{} (\rslcpp{}) implements this approach, enabling existing nodes to be combined into a simulation routine that yields reproducible results usually without requiring any source code changes.
    We demonstrate that our approach produces identical results across various CPUs and architectures when testing both a synthetic benchmark and a real-world robotics system. \rslcpp{} is open\nobreakdash-sourced at \url{https://github.com/TUMFTM/rslcpp}.
\end{abstract} %
\section{Introduction}
\begin{figure}[!t]
    \centering
    \vspace*{0.2cm}
    \begin{subfigurecustom}
        \begin{minipage}{\columnwidth}
            \centering
            \startsubfigcustom
            \inputtikzfig{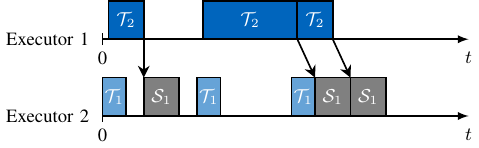}
            \par\vspace{-0.5mm}
            \subfigcaptioncustom{Asynchronous multi-process execution}
            \label{fig:asynchronous_execution}
            \vspace*{0.5cm}
        \end{minipage}
        \begin{minipage}{\columnwidth}
            \centering
            \startsubfigcustom
            \inputtikzfig{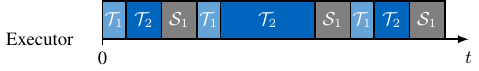}
            \par\vspace{-0.5mm}
            \subfigcaptioncustom{Deterministic synchronous execution using our approach.}
            \label{fig:synchronous_execution}
        \end{minipage}
    \end{subfigurecustom}
    \caption{Illustrative comparison of asynchronous and synchronous execution. The example consists of timer~(\schedChartlabelTone,~\schedChartlabelTtwo) and subscription~(\schedChartlabelSone) callbacks. \schedChartlabelSone{} is triggered by a publisher at the end of \schedChartlabelTtwo. In the asynchronous case, the callback runtimes influence the order in which the callbacks are processed. With our approach, they remain in a fixed order. $t$ indicates wall time progression, which is decoupled from the simulation time for the synchronous example.}
    \label{fig:first_page_figure}
\end{figure}

Modern robotics is a rapidly growing field that has garnered significant attention from both academia and industry in recent years~\cite{Seyed2024, haghighi2025comprehensivereviewbibliometric}. The \softwarefull{Robot Operating System} (\roswithouttwo{})~\cite{quigley2009, macenski2022RobotOperatingSystem} is a popular middleware and software framework that enables the development and deployment of modular robotic systems. It has been widely adopted across a broad range of robotic applications, including both commercial products and research, spanning wheeled and legged ground robots, aerial drones, and underwater vehicles~\cite{macenski2022RobotOperatingSystem, macenski2023}. Simulation plays a pivotal role in these fields, as it enables the evaluation and development of algorithms without the need for expensive hardware, thereby allowing a broader community to contribute to research~\cite{Seyed2024}.
Although \ros{} provides mechanisms for simulated time, callback execution order cannot be guaranteed due to its asynchronous, multi-process design~\cite{macenski2022RobotOperatingSystem}. Figure~\ref{fig:first_page_figure} illustrates that computation times, transmission delays, and scheduling overhead can significantly influence the execution order of callbacks in such a multi-process system.
However, both the execution order and the timing of these callbacks can have a substantial impact on the outcomes of complex multi-process simulations~\cite{betz2023AnalysisSoftwareLatency,casini2019ResponseTimeAnalysisROS}.
As a consequence, achieving exact reproducibility of simulation results remains challenging, despite an identical software setup. This issue is further exacerbated when testing on various hardware platforms, such as during continuous integration testing.
As a workaround, code is frequently benchmarked outside its \ros{}-based implementation~\cite{vizzo2023KISSICPDefensePointtoPoint}.
However, as the integration into \ros{} often constitutes a significant portion of the codebase~\cite{sagmeister2025ApproachingCurrentChallenges}, testing in isolation fails to validate the complete system and neglects potential side effects.

To address the aforementioned limitations, we introduce a methodology for achieving deterministic execution order and timing of callbacks in a multi-node \ros{} system.
By decoupling simulation progress from computation time, our approach enables sped-up simulations or the evaluation of algorithms that are not yet real-time capable.
This acts as a robust baseline for scientific benchmarking and continuous integration testing.
For this, the main contributions of this paper are:
\begin{itemize}
    \item We present a methodology for enforcing deterministic execution of \ros{} callbacks, thereby enabling reproducible simulations across varying hardware platforms.
    \item We validate the proposed approach across multiple CPU architectures using both a synthetic benchmark and a real-world robotic system.
    \item We implement and open-source the presented methodology in the \rslcppfull{} (\rslcpp). This library enables the creation of deterministic simulations using existing \ros{} nodes usually without requiring any source code changes.
\end{itemize} %
\section{Related Work}

Only a few approaches in the literature address the lack of predictability in \roswithouttwo{}-based systems.
Otto~\cite{otto2023EnablingReproducibilityROS} identifies different sources of non-deterministic execution in \ros{} and proposes a method to ensure reproducible execution orders.
This work introduces an orchestrator that uses callback graphs and ordering constraints to schedule the execution of callbacks in a deterministic way.
However, this orchestrator requires a specific callback structure, hence necessitating modifications to the code of many \ros{} nodes.
Other works~\cite{choi2021PriorityDrivenChainAwareScheduling, randolph2021ImprovingPredictabilityInRos} propose \ros{} executors that aim to improve the predictability of callback execution.
However, their focus lies on real-time capability rather than determinism.
These works assign priorities to each callback or chain of callbacks based on real-time requirements.
To ensure execution according to the assigned priorities, Randolph~\cite{randolph2021ImprovingPredictabilityInRos} utilizes a multi-threaded executor.
In contrast, Choi et al.~\cite{choi2021PriorityDrivenChainAwareScheduling} employ multiple executors for different callback chains, assigning each executor to a single core.

Furthermore, the \gazebo{} simulator~\cite{koenig2004Gazebo} offers a lockstep mode.
This mode allows the executor to pause the simulation, enabling synchronization of all sensor outputs.
However, the execution order of the remaining software interacting with the simulator remains uncontrolled and can still lead to non-deterministic behavior.

Additionally, deterministic executors for \microros{}, a \ros{} variant for microcontrollers, have been proposed.
Staschulat et al.~\cite{staschulat2020rclcExecutorDomainspecific} present a deterministic \rclc{} executor that can execute callbacks in a user-defined order and handle various trigger conditions for callback processing.
In a later work, this executor has been extended to a budget-based multi-threaded executor to improve predictability in real-time applications~\cite{staschulat2021budget}.
Similarly, Wang et al.~\cite{wang2023tide} extend the deterministic \rclc{} executor to address the problem of priority inversion, where lower-priority tasks block high-priority tasks.
They propose an executor with a list of ready tasks sorted by priority to replace the set in the original executor, ensuring the earlier execution of high-priority tasks.

In summary, various approaches exist to improve callback scheduling and execution order. However, none provide a comprehensive solution for creating deterministic simulations that can be easily integrated without requiring heavy modification of the existing \ros{} codebase. %
\section{Methodology}

To create a deterministic simulation using \ros{} nodes, we first discuss the foundations for achieving deterministic callback execution. Based on these concepts, we present our approach for efficiently specifying simulation routines.

\subsection{Achieving Deterministic Callback Execution}

To achieve reproducible simulation results in a generic \ros{} system, all callbacks must be executed in a defined order with precise timing.
Since this is challenging in an asynchronous multi-process system like \ros{}, our approach first aggregates all nodes into a single process by utilizing node composition.
We require all simulation components to reside within this single process, isolated from external communication.
This simplifies scheduling, as all callbacks run within a single thread, and enables efficient zero-copy intra-process communication~\cite{macenski2023ImpactROS2}.
While single-process composition is fundamental to our approach, it has drawbacks such as precluding fault isolation typical in real-world deployments. However, these constraints are generally acceptable in the context of simulation and testing.

\begin{figure}[!t]
    \vspace*{0.2cm}
    \begin{center}
        \inputtikzfig{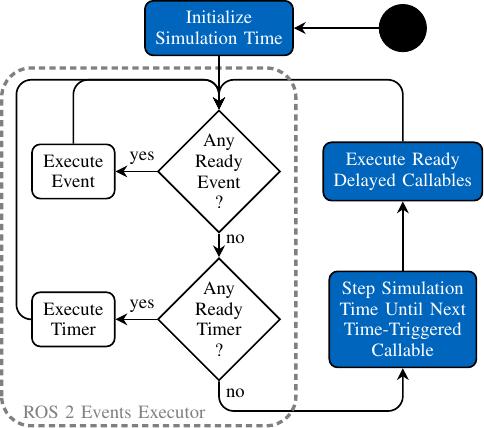}
    \end{center}
    \par\vspace*{-0.35cm}
    \caption{\rslcpp{} event loop. The actions depicted in blue are specific to the \rslcpp{} event loop, while white coloring indicates processes implemented within the \rclcpp{} events executor. The loop terminates before advancing time as soon as the job object indicates that the simulation is complete.}
    \par\vspace{-2mm}
    \label{fig:method_overview}
\end{figure}

To establish deterministic execution, we implement a custom event loop, as illustrated in Fig.~\ref{fig:method_overview}.
Within this event loop, we utilize the \rclcpp{} events executor first introduced in \ros{} Jazzy. This executor processes callbacks in a defined first-in-first-out manner~\cite{teper2024BridgingGapROS2, teper2025ReconcilingROS2}.
To become independent of computation times, we configure all nodes to run in simulation time.
We advance simulation time only after the executor has finished executing all ready entities. This includes both time-triggered callbacks (e.g., timers) and successive data-triggered callbacks (e.g., subscriptions, services, clients).
However, this scheme requires that the system eventually becomes idle if time is not advanced.
Consequently, an endless cycle of instantaneous data-triggered callbacks prevents the executor from ever becoming idle, halting the progression of simulation time.
Nonetheless, this limitation rarely affects real-world robotic systems, as they commonly rely on periodic control and sensor loops. Furthermore, the delay mechanism described below can mitigate this issue.

Since the entire simulation runs in a single thread, a blocking callback (e.g., waiting for a service response) will halt the entire simulation. While this requires careful implementation of callbacks, it significantly simplifies debugging, as a breakpoint in any callback effectively pauses the entire system state.
To achieve maximum efficiency, we advance simulation time adaptively to the next time-triggered callback, effectively creating a discrete-event simulation~\cite{robinson2004Simulationpracticemodel}.
\subsection{Delay Modeling}

Since the discrete-event approach executes callbacks without advancing simulation time, it neglects the impact of computation time on system latency. To address this, we introduce a mechanism to artificially delay message publication. %
Instead of publishing a message immediately upon generation, we schedule its publication to occur after a configurable duration.
Publishing the delayed message is handled by the custom event loop itself (Fig.~\ref{fig:method_overview}).

\subsection{Simulation Specification and Integration}

To specify generic simulation routines, we propose a job object that encapsulates all necessary information for executing a simulation.
It specifies the start timestamp and defines which nodes are part of the simulation.

Since \ros{} allows setting parameters at launch, we utilize its dynamic parameter loading instead of specifying parameters within the job itself.
Therefore, the same job object can conduct various simulations by loading different parameters.
Additionally, the job object terminates the event loop as soon as the simulation routine is completed. For this, the job object provides a boolean flag that indicates whether the simulation has finished. Further, the job object provides an exit code indicating successful or failed execution.

To simplify the integration of existing \ros{} nodes into \rslcpp{} simulations, we provide a dynamic job executable. This enables loading nodes dynamically at launch by utilizing the \rclcpp{} component system~\cite{macenski2023ImpactROS2}.
Consequently, the full simulation routine can be configured exclusively at launch.
As a result, existing \rclcpp{} components can be used for deterministic simulations, usually without modifying their source code. However, recompiling node components against our modified \rclcpp{} version is required for the system to function correctly.
Given that source code can contain arbitrarily complex constructs, we cannot formally guarantee that every conceivable node will work without altering its source code. However, we successfully executed all nodes tested in our evaluation without requiring any source code modifications. %

\section{Experimental Setup}

We conduct two different validation experiments. The first experiment focuses on a synthetic benchmark system designed to test the determinism of callback execution. The second experiment uses a LiDAR odometry benchmark to demonstrate the applicability of our approach in a real-world robotics scenario.

\subsection{Synthetic Benchmark System}

\begin{figure}[!t]
    \centering
    \vspace*{0.2cm}
    \inputtikzfig{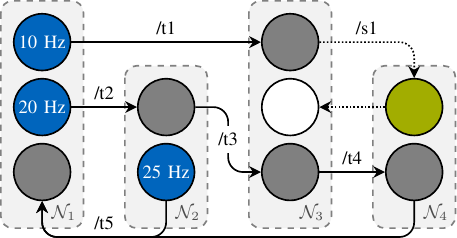}
    \caption{Synthetic benchmark system for testing deterministic callback execution. Timer callbacks with their respective frequencies are shown in blue, while subscription callbacks are depicted in dark gray. Service callbacks are colored green while a client's response callback is depicted in white. Individual nodes are represented as light gray boxes.}
    \label{fig:synthetic-benchmark-system}
\end{figure}

To experimentally validate the determinism of our approach, a system that is highly sensitive to callback execution order and timing is required. To achieve this, we designed a synthetic benchmark system, which is depicted in Fig.~\ref{fig:synthetic-benchmark-system}.

It consists of multiple nodes interconnected via publishers, subscribers, services, and clients.
The system is designed to ensure that multiple callbacks (subscriptions, timers)
become ready simultaneously, both within a single node and across multiple nodes.
This increases the potential for execution order variance when the system is executed conventionally.

To make the system sensitive to execution order and timing, each node maintains an internal state represented by a hash value. Additionally, each callback has a unique, randomly generated ID.
Each time a callback is executed, it modifies the internal state of the node using a pseudo-random number generator~\cite{steele2014Fastsplittablepseudorandom}.
This random number generator is fed with the callback's ID, the execution timestamp, the node's current internal state, and, if available, the callback's input data.
Additionally, the nodes are interconnected by publishing their constantly updated internal state.%

Comparing the system's final state after a fixed simulation period across runs allows us to detect any non-determinism in callback execution.

To demonstrate that our approach is also invariant to fluctuations in computation times, each callback randomly sleeps between \num{0} and \SI{20}{\milli\second} before executing its logic.

\subsection{Case Study: LiDAR Odometry}
To demonstrate the applicability of our framework to real-world robotics, we benchmark the LiDAR odometry algorithm \kissicp{}~\cite{vizzo2023KISSICPDefensePointtoPoint} using the \kittiodometry{}~\cite{KITTI} Sequence 00.
\kissicp{} was selected for its algorithmic simplicity, which facilitates highly reproducible results when executed in a deterministic environment.

The simulation setup comprises three \ros{} nodes: a rosbag player, the \kissicp{} node, and a data recorder.
The rosbag player publishes the \kitti{} data, which was pre-converted to a rosbag using \kittitobag{}~\cite{krejci2019kitti2bag}.
The recorder logs all topics to a new rosbag, which is subsequently used to compute the root-mean-squared error (RMSE) of the average positional error (APE) via \evo{}~\cite{grupp2017evo}.

For each CPU, we constrain \kissicp{}'s number of parallel threads to the minimum required for the algorithm to reliably converge on the whole scenario. This simulates a resource-constrained robotics system while accommodating processors with limited single-core performance.
Additionally, limiting the subscriber queue history depth to 1 and the number of ICP iterations to 100 ensures the low latency necessary for real-world applications. %
\section{Results}

\subsection{Synthetic Benchmark System}
We ran the synthetic benchmark on various architectures and processors to demonstrate the deterministic execution of our approach. Table~\ref{tab:determinism_validation} lists the evaluated CPU models.
For each system, we performed 100 independent runs and compared the final hash values produced by all nodes.
Across all 600 simulations, the final hash values were identical, confirming that the simulation framework yields fully deterministic, bit-identical results under the conditions tested.

\begin{table}[!t]
    \vspace*{0.2cm}
    \caption{CPU models used for validating deterministic execution with the synthetic benchmark system.}
    \label{tab:determinism_validation}
    \centering
    \begin{tabular}{c c c c}
        \toprule
        \textbf{Architecture} & \textbf{Vendor} & \textbf{Model}    & \textbf{Cores/Threads} \\
        \midrule
        x86\_64               & Intel           & i7-11850H         & 8/16                   \\
        x86\_64               & AMD             & Ryzen 7 Pro 6850U & 8/16                   \\
        x86\_64               & Intel           & Xeon D-2166NT     & 12/12                  \\
        x86\_64               & AMD             & EPYC 7313P        & 16/32                  \\
        x86\_64               & AMD             & Ryzen 9 7950X     & 16/32                  \\
        aarch64               & ARM             & Cortex-A72        & 4/4                    \\
        \bottomrule
    \end{tabular}
    \vspace{-0.8mm}
\end{table}

\subsection{Delay Modeling}
An empirical test available in our repository demonstrates that our approach is able
to achieve nanosecond-precise synthetic delays while preserving determinism.
Therefore, the delay mechanism allows for the analysis of system performance under varying latency conditions without requiring runtime optimization of the underlying algorithms.

\subsection{Case Study: LiDAR Odometry}
\begin{figure}[!t]
    \vspace*{0.2cm}
    \centering
    \begin{subfigurecustom}
        \begin{minipage}[t]{0.49\columnwidth}
            \centering
            \startsubfigcustom
            \inputtikzplot{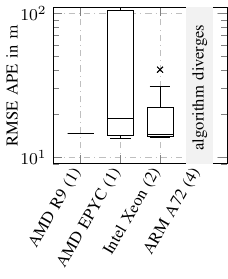}
            \vspace{-0.35cm}
            \subfigcaptioncustom{\ros}
            \label{fig:performance_benchmark_results_ros2}
        \end{minipage}
        \begin{minipage}[t]{0.49\columnwidth}
            \centering
            \startsubfigcustom
            \inputtikzplot{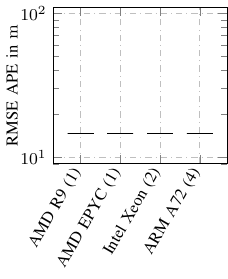}
            \vspace{-0.35cm}
            \subfigcaptioncustom{\rslcpp}
            \label{fig:performance_benchmark_results_rslcpp}
        \end{minipage}
    \end{subfigurecustom}
    \par\vspace*{-0.1cm}

    \caption{Root-mean-squared-error (RMSE) of the average positional error (APE) over 100 simulations of the \kissicpbreaking{}~\cite{vizzo2023KISSICPDefensePointtoPoint} on Sequence~00 of the \kittiodometry{}~\cite{KITTI} dataset within \ros{} and \rslcpp. The number of parallel \kissicpbreaking{} threads is reported in parentheses. CPU models are abbreviated and correspond to Table~\ref{tab:determinism_validation}.}
    \label{fig:performance_benchmark_results}
    \vspace{-2mm}
\end{figure}

Fig.~\ref{fig:performance_benchmark_results} presents the results from running the LiDAR odometry benchmark using both native \ros{} and our framework.

Experiments conducted natively in \ros{} exhibit substantial variability across the various hardware architectures (Fig.~\ref{fig:performance_benchmark_results_ros2}).
While the AMD Ryzen 9 7950X exhibits fully reproducible behavior, experiments on the AMD EPYC 7313P and Intel Xeon D-2166NT show significant result jitter.
The \kissicp{} diverges on the ARM Cortex-A72 because the CPU's computational resources are inadequate to maintain a sufficient update rate.

In contrast, Fig.~\ref{fig:performance_benchmark_results_rslcpp} indicates that even the multi-threaded implementation of \kissicp{} produces repeatable results when executed in \rslcpp.
Similar to the results obtained from the synthetic benchmark system, all runs across all hardware architectures tested yield identical results.

\section{Discussion and Conclusion}
\label{sec:conclusion}

In this work, we introduced a methodology for creating deterministic simulations by aggregating \ros{} nodes into a single process controlled by a custom event loop.
Our experiments demonstrated bit-exact reproducibility across various systems, eliminating the variance typical of asynchronous multi-process execution.

It is important to note, however, that the validation of determinism is empirical rather than mathematically proven, as the complexity of \ros{} and its client libraries complicates formal verification.
By utilizing the existing component system and executor architecture, our framework is tightly integrated with \ros{}, simplifying the adoption for existing \ros{} projects.
However, this tight coupling also introduces a dependency on the underlying executor logic of \ros{}, which may necessitate strict monitoring of determinism as well as maintenance efforts to ensure compatibility with future \ros{} releases.

A further limitation of our approach is that nodes must finish their callbacks in finite time and must not engage in cyclic undelayed publish-subscribe loops to avoid stalling the executor. While single-threaded execution itself processes callbacks sequentially, thereby increasing the runtime of a single complex simulation, our deterministic framework enables the execution of multiple simulations in parallel. This potentially even increases throughput. Furthermore, internal tests with complex multi-node systems indicated no scalability issues.

In conclusion, \rslcpp{} offers a modular, open-source solution for reproducible robotic simulation across different hardware setups.
Rather than improving determinism in real-world deployments, \rslcpp{} is tailored for scientific research, system-level debugging, and continuous integration testing. 
\section*{Acknowledgment}
Author contributions:
S. Sagmeister, as the first author, designed the article's structure and essentially contributed to developing and implementing the framework.
M. Weinmann and P. Pitschi contributed to the writing, design, and implementation.
M. Lienkamp made an essential contribution to the concept of the research project. He revised the paper critically for important intellectual content. M. Lienkamp gives final approval for the version to be published and agrees to all aspects of the work. As a guarantor, he accepts responsibility for the overall integrity of the paper.
AI tools (GPT5.2, Gemini 3 Pro) were used exclusively for editorial, code, and documentation refinement of author-provided input.

\end{document}